# Reachable Distance Function for KNN Classification

Shichao Zhang , *Senior Member, IEEE*, Jiaye Li , and Yangding Li

**Abstract**—Distance function is a main metrics of measuring the affinity between two data points in machine learning. Extant distance functions often provide unreachable distance values in real applications. This can lead to incorrect measure of the affinity between data points. This paper proposes a reachable distance function for KNN classification. The reachable distance function is not a geometric direct-line distance between two data points. It gives a consideration to the class attribute of a training dataset when measuring the affinity between data points. Concretely speaking, the reachable distance between data points includes their class center distance and real distance. Its shape looks like "Z," and we also call it a Z distance function. In this way, the affinity between data points in the same class is always stronger than that in different classes. Or, the intraclass data points are always closer than those interclass data points. We evaluated the reachable distance with experiments, and demonstrated that the proposed distance function achieved better performance in KNN classification.

**Index Terms**—Distance functions, reachable distance, machine learning, KNN classification

✦

## 1 INTRODUCTION

In machine learning applications, we must measure the affinity between data points, so as to carry out data clustering, or data classification. A natural way of measuring the affinity between data points is euclidean distance function, or its variants. It is true that euclidean distance is geometrically the shortest distance between two data points, which is only a spatially direct-line measure [1]. However, euclidean distance function and its variants often provide unreachable distance values in real applications. This can lead to incorrect measure of the affinity between data points. In other words, this is not a real reachable distance in real applications although euclidean distance function and its variants have been widely adopted in data analysis and processing applications. We illustrate this with two cases as follows.

*Case I.* There is a gap, such as a frontier, a river/sea, and a mountain, between two points, see Fig. 1. This gap can often be unbridgeable. In other words, these two points are not able to reach in euclidean distance time from each other.

*Case II.* Two data points are packed in different bags/shelves, illustrated in Fig. 2. For example, in a clinic doctors always put medical records of benign tumors into a bag, and all medical records of malignant tumors into another bag. This indicates that a medical record concerning a benign

---

- *Shichao Zhang and Jiaye Li are with the School of Computer Science and Engineering, Central South University, Changsha 410083, China. E-mail: {zhangsc, lijiaye}@csu.edu.cn.*
- *Yangding Li is with the Hunan Provincial Key Laboratory of Intelligent Computing and Language Information Processing, Hunan Normal University, Changsha 410081, China. E-mail: lyd271@126.com.*

*Manuscript received 17 Mar. 2021; revised 13 May 2022; accepted 18 June 2022. Date of publication 0 . 2022; date of current version 0 . 2022. This work was supported in part by the Key Program of the National Natural Science Foundation of China under Grant 61836016. (Corresponding authors: Jiaye Li and Yangding Li.) Recommended for acceptance by E. Chen. Digital Object Identifier no. 10.1109/TKDE.2022.3185149*

tumor is very far from any medical records of malignant tumors.

From the above Cases I and II, the euclidean distance between two data points cannot be the reachable distance. In other words, reachable distance between two data points has been an open problem. However, in Big Data mining, we must measure the affinity between data points with proper metrics. These motivate us to well understand training data and design a reachable distance function.

In real data analysis and processing applications [2], ones often directly apply euclidean distance function and its variants to measure the affinity between data points without understanding training data. This can lead to incorrect measure of the affinity between data points. And the discovered results do not meet real applications. It is true that data collectors always ignore some information, such as that in the above two cases in data preparation stage (data collection). For example, when medical records are collected and input to the computer systems in a clinic, data collector often takes all the medical records out of from their bags without distinguishing them unlike doctors. After stored in the systems, these medical records look like coming from the same bag, and the natural separation information has passed away. To make data analysis algorithms applicable, the euclidean distance function or its variants are employed to measure the affinity between data points, whether the euclidean distance between in two data points is the reachable distance or not.

While the natural separation and unreachable information are missed in data collection stage, training data should be well understood before mining them. Without other information, in this paper we advocate to take the class center distance as the natural separation information, and design a reachable distance function for KNN classification. The reachable distance between data points includes their class center distance and real distance. And its shape looks like "Z," it is also called as Z distance function. If their class center distance is large enough, the affinity of two data

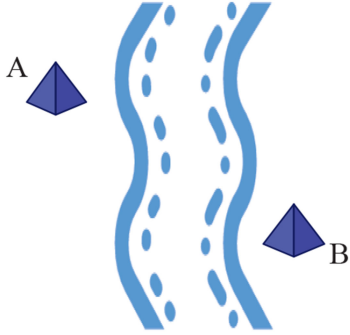

Fig. 1. There is a river between A and B.

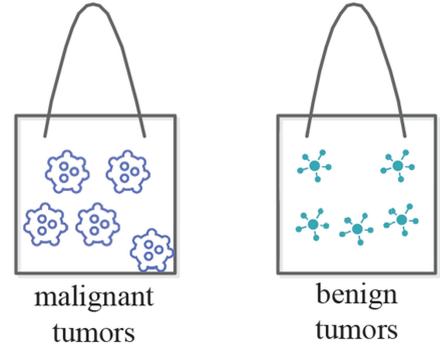

malignant tumors

benign tumors

Fig. 2. Medical records with different classes are often packed in different bags.

points in the same class can always be stronger than that in different classes. In this way, the intraclass data points are always closer than those interclass data points in a training dataset. The reachable distance is evaluated with experiments, and demonstrated that the proposed distance function achieved better performance in KNN classification.

The rest of this paper is organized as follows. Related work and some concepts are recalled in Section 2. The Z distance (reachable distance) is proposed in Section 3. The reachable distance is evaluated with experiments in Section 4. This paper is concluded in Section 5.

## 2 RELATED WORK

This section first recalls traditional distance functions. And then, some new distance functions are briefly discussed. Finally, it simply discussed that euclidean distance is often not the reachable distance in real applications.

### 2.1 Traditional Distance Functions

The demand of distance function is everywhere in real life [3]. For example, when building a railway, we must roughly calculate the required construction materials according to the distance between the two places [4]. The arrival time of express delivery is often related to the distance [5]. High jumpers must take off within the prescribed distance [6]. The height of basketball frame must be unified by height calculation [7]. This shows that the distance function is very important in social life [8]. Common distance functions include euclidean distance [9], Manhattan distance [10], Chebyshev distance [11], standardized euclidean distance [12], Mahalanobis distance [13], Bhattacharyya distance [14], Kullback-Leibler divergence [15], Hamming distance [16] and cosine distance [17]. Next, we introduce these distance functions in detail.

Traditional distance function is defined as follows.

$$d(\mathbf{a}, \mathbf{b}) = \left[ \sum_{j=1}^{d} |a_j - b_j|^p \right]^{\frac{1}{p}}, \quad (1)$$

where $\mathbf{a}$ and $\mathbf{b}$ are two sample points and $d$ is the dimension of each sample (i.e., the number of features). In Eq. (1), when $p = 1$, $d(\mathbf{a}, \mathbf{b})$ is the Manhattan distance, when $p = 2$, $d(\mathbf{a}, \mathbf{b})$ is the classical euclidean distance, and when $p = infinity$, $d(\mathbf{a}, \mathbf{b})$ is the Chebyshev distance. These three kinds of distance are the most common distance measures. euclidean distance is a very intuitive distance measure, which has a wide range of applications. However, it is not suitable for high-dimensional data. Manhattan distance is called city block distance, which is more non intuitive than euclidean distance, and it is not the shortest path. Chebyshev distance is the maximum distance along the coordinate axis, which can only be applied to specific situations.

In view of the different distribution of each dimension in the data, the standardized euclidean distance improves the euclidean distance, i.e., each feature is standardized to have the same mean variance, as shown in the following formula.

$$\tilde{\mathbf{X}} = \frac{\mathbf{X} - \mu}{\sigma}. \quad (2)$$

Each point in $\mathbf{X}$ is normalized by the Eq. (2), $\mu$ is the mean, and $\sigma$ is the variance. $\tilde{X}$ is the data set after standardization. The standardized euclidean distance is as follows:

$$d(\mathbf{a}, \mathbf{b}) = \sqrt{\sum_{j=1}^{d} \left( \frac{a_j - b_j}{\sigma_j} \right)^2}. \quad (3)$$

From the Eq. (3), the standardized euclidean distance adds $1/\sigma_j$ to the euclidean distance, which can be regarded as a weighted euclidean distance.

Mahalanobis distance is also a variant of euclidean distance. The Mahalanobis distance is defined as follows.

$$d_M(\mathbf{a}, \mathbf{b}) = \sqrt{(\mathbf{a} - \mathbf{b})\mathbf{S}^{-1}(\mathbf{a} - \mathbf{b})^T}, \quad (4)$$

where $\mathbf{S}$ is the covariance matrix. It can be seen from Eq. (4) that if the covariance matrix is a identity matrix, the Mahalanobis distance is the same as the euclidean distance. If the covariance matrix is a diagonal matrix, then the Mahalanobis distance is the same as the standardized euclidean distance. It should be noted that Mahalanobis distance requires that the number of samples of data is greater than the number of dimensions, so that the inverse matrix of covariance matrix $\mathbf{S}$ exists. Its disadvantage is computational instability due to covariance matrix $\mathbf{S}$.

The Bhattacharyya distance is a measure of the similarity between two probability distributions, as shown in the following formula.

$$d_B(p, q) = -\ln(BC(p, q)), \quad (5)$$

where $p$ and $q$ are the two probability distributions on data $\mathbf{X}$ respectively. If it is a discrete probability distribution,

then $BC(p,q) = \sum_{x \in X} \sqrt{p(x)q(x)}$. If it is a continuous probability distribution, then $BC(p,q) = \int \sqrt{p(x)q(x)}dx$. The KL (Kullback-Leibler) divergence is similar to the Bhattacharyya distance. It can also measure the distance or similarity of two probability distributions. As shown in the following formula:

$$KL(p||q) = \sum p(x)\log \frac{p(x)}{q(x)} \tag{6}$$

$$KL(p||q) = \int p(x)\log \frac{p(x)}{q(x)}dx. \tag{7}$$

Eqs. (6) and (7) are the discrete probability distribution and continuous probability distribution, respectively. KL divergence has a wider range of applications relative to Mahalanobis distance.

In data transmission error control coding, Hamming distance is often used to measure the distance between two characters. It describes the number of different values in the two codes. The formula is defined as follows.

$$d_H(\mathbf{a}, \mathbf{b}) = \sum_{j=1}^{d} a_j \oplus b_j, \tag{8}$$

where $\oplus$ is the XOR operation. Both $\mathbf{a}$ and $\mathbf{b}$ are $n$-bit codes. For example: $\mathbf{a} = 11100111$, $\mathbf{b} = 10011001$, then the Hamming distance between $\mathbf{a}$ and $\mathbf{b}$ is $d(\mathbf{a}, \mathbf{b}) = 6$. Hamming distance is mostly used in signal processing. It can be used to calculate the minimum operation required from one signal to another.

In addition to the above distance functions, there is also a cosine distance function. It is derived from the calculation of the cosine of the included angle, as shown in the following formula.

$$d_C(\mathbf{a}, \mathbf{b}) = 1 - \frac{\mathbf{a} \cdot \mathbf{b}}{\|\mathbf{a}\|\|\mathbf{b}\|} = 1 - \frac{\sum_{j=1}^{d} a_j b_j}{\sqrt{\sum_{j=1}^{d} a_j^2}\sqrt{\sum_{j=1}^{d} b_j^2}}. \tag{9}$$

Cosine distance is mostly used in machine learning algorithms to calculate the distance or similarity between two data points. Its value range is [0, 2], which satisfies the non-negativity of the distance function. Its disadvantage is that it only considers the direction of two samples, and does not consider the size of their values.

## 2.2 Different Distance Measures for KNN Classification

The above distance functions have their own characteristics and applicable scopes, i.e., they are developed for different application requirements. e.g., euclidean distance is suitable for low dimensional data. Its performance will decline with the increase of dimension. The cosine distance can be well applied to high-dimensional data. Hamming distance is suitable for error correction and detection when transmitting data on computer network.

In addition to the KNN based on the traditional distance function, researchers have also proposed some new distance functions for KNN [18]. Gou et al. proposed a distance function for KNN based on the local mean vector [19]. Specifically, it first finds K nearest neighbors in each class, and uses these neighbors to construct a local mean vector, and each class constructs K local mean vectors. Then it calculates the distance between the test sample and each local mean vector in each class. Finally, it selects the class of the local mean with the smallest distance as the predicted class of the test data. Poorheravi et al. proposed a triple learning method to perform metric learning [20]. It not only uses hierarchical sampling to build a new triple mining technology, but also analyzes the proposed method on three public data sets. Song et al. proposed a parameter-free metric learning method [21]. This method is a supervised metric learning algorithm. Specifically, it discards the cost term, so that there is no need to set the parameters required to adjust the validation set. In addition, it only considers recent imposters, which greatly reduces time costs. In the experiment, it has achieved better results than the traditional nearest neighbor algorithm with large margin. Noh et al. proposed a local metric learning for nearest neighbor classification [22]. It uses the deviation caused by the limited sampling effect to find a suitable local metric, which can reduce the deviation. In addition, it also applies the dimensionality reduction theory to metric learning, which can reduce the time cost of the algorithm. Ying et al. proposed a semi-supervised metric learning method [23]. Specifically, it first uses the structural information of the data to formulate a semi-supervised distance metric learning model. Then it transforms the proposed method into a problem of minimizing symmetric positive definite matrices. Finally, it proposes an accelerated solution method to keep the matrix symmetric and positive in each iteration. Wang et al. proposed a robust metric learning method [24]. This method is an improvement of the nearest neighbor classification with large margin. Its main idea is to use random distribution to estimate the posterior distribution of the transformation matrix. It can reduce the influence of noise in the data, and the anti-noise of the algorithm is verified in experiments. Jiao et al. proposed a KNN classification method based on pairwise distance metric [25]. It uses the theory of confidence function to decompose it into paired distance functions. Then it is adaptively designed as a pair of KNN sub-classifiers. Finally, it performs multi-classification by integrating these sub-classifiers. Song et al. proposed a high-dimensional KNN search algorithm through the Bregman distance [26]. Specifically, it first partitions the total dimensions to obtain multiple subspaces. Then it gets the effective boundary from each partition. Finally, it uses ensemble learning to gather the various partitions. Su et al. learn the meta-distance of a sequence from virtual sequence regression [27]. The meta-distance obtained by the ground measurement makes the sequences of the same category produce smaller values, and the sequences of different categories produce larger values. In addition, it also verified the effectiveness of the proposed algorithm on multiple sequence data sets. Faruk Ertugrul et al. proposed a new distance function [28]. It uses differential evolution method to optimize parameters based on metadata, and applies the proposed distance function to KNN. In addition, it also verified the performance of the algorithm on 30 public data sets. In order to facilitate readers to understand the differences between various distance functions, we summarize them from the measurement affinity, class information, parameters and reachable distance, as shown in Table 1.

TABLE 1
Comparison of Various Distance Functions in Related Work

| Distance Function | Measuring Affinity | Class Information | With or Without Parameters | Reachable Distance |
|---|---|---|---|---|
| Euclidean distance | ✓ | × | × | × |
| Manhattan distance | ✓ | × | × | × |
| Chebyshev distance | ✓ | × | × | × |
| Mahalanobis distance | ✓ | × | × | × |
| Bhattacharyya distance | ✓ | × | × | × |
| Kullback-Leibler divergence | ✓ | × | × | × |
| Hamming distance | ✓ | × | × | × |
| Cosine distance | ✓ | × | × | × |
| Gou et al. [19] | ✓ | ✓ | ✓ | × |
| Poorheravi et al. [20] | ✓ | ✓ | ✓ | × |
| Song et al. [21] | ✓ | ✓ | × | × |
| Noh et al. [22] | ✓ | ✓ | ✓ | × |
| Ying et al. [23] | ✓ | ✓ | ✓ | × |
| Wang et al. [24] | ✓ | ✓ | ✓ | × |
| Jiao et al. [25] | ✓ | ✓ | ✓ | × |
| Song et al. [26] | ✓ | × | ✓ | × |
| Su et al. [27] | ✓ | ✓ | ✓ | × |
| Faruk Ertugrul et al. [28] | ✓ | × | ✓ | × |

Researchers also compare the performance of different distance functions. e.g., Geler et al. measured the impact of each elastic distance on the weighted KNN classification in time series data [29]. In the experiment, it lists the values of each parameter in detail, compares different elastic distances and verifies that the weighted KNN always outperforms 1NN in time series classification. Feng et al. analyzed the performance of the KNN algorithm according to different distance functions, including Chebyshev distance, euclidean distance and Manhattan distance and cosine distance [30]. In addition, it also compares the performance of some new distance functions. In KNN classification, most of the performance of the new distance function is better than euclidean distance. Although these new distance functions improve KNN from different views, they all ignore the accessibility of distance, i.e., the natural distance in the data is not found. In addition to improving KNN through distance function, some researchers also improve KNN for specific problems, such as Kiyak et al. proposed a new KNN algorithm for multi-view data [31]. Specifically, it first constructs a weak classifier under each view. Then the weak classifiers in the previous step are integrated to form a multi-view classification model. Finally, the performance of the algorithm is verified on multi-view data sets. It can be regarded as an integrated learning algorithm, i.e., it only divides multi-view data into multiple single views for multiple learning, without considering the relationship between views. Maillo et al. proposed a fast and scalable fuzzy k-nearest neighbor algorithm [32]. It uses local hybrid spill tree to divide the data set into multiple subsets, and calculates the class membership degree of each subset. In addition, it also uses the global approximate hybrid spill tree to generate a tree from the training data, so as to consider the class membership degree of all samples. In this way, the method considers not only the local structure information of the data, but also the global structure of the data.

Distance function can be used not only for KNN, but also for missing value filling, class imbalance classification, multi label learning and clustering. It has a wide range of applications, e.g., Seoane Santos et al. used KNN to perform missing value interpolation through different distance functions, and verified the effects of different distance functions [33]. Marchang et al. used KNN to propose a sparse population perception model [34]. It considers spatial correlation and temporal correlation in the algorithm respectively. In addition, the correlation between time and space is also embedded in the proposed method. Experiments have also shown that KNN, which considers the correlation between time and space, has a better effect in the inference of missing data. Valverde, et al. used KNN for text classification, and carried out the influence of different distance functions on text classification [35]. Susan and Kumar proposed a combination of metric learning and KNN for class imbalance data classification [36]. Specifically, it first performs spatial transformation on the data. Then it divides the K test samples into two clusters according to the distance of the two extreme neighbors. Finally, the majority vote rule is used to determine the class label of test data. Although these researchers have proposed some new measurement functions, none of them really takes the natural distance into account in the data. Sun et al. proposed a metric learning for multi-label classification [37]. It is modeled by the interaction between the sample space and the label space. Specifically, it first adopts matrix weighted representation based on component basis. Then it uses triples to optimize the weight of the components. Finally, the effectiveness of the combined metric in multi-label classification is verified on 16 benchmark data sets. Gu et al. proposed a new distance metric for clustering [38]. This method combines the advantages of euclidean distance and cosine distance. It can be applied to clustering to solve high-dimensional problems. Gong et al. used indexable distance to perform nearest neighbor query [39]. It uses kd-tree to further improve the search speed of the algorithm. Wang analyzed the multimodal data and showed the importance of distance function in depth multimodal method [40]. Wang et al. proposed a dimensionality reduction algorithm for multi view data, which is called kernel multi view subspace analysis [41]. It uses self weighted learning to apply appropriate weights to different views. After reducing the dimension of multi view data, this method can greatly increase the applicability of distance function.

In addition to the above application of improving KNN based on distance function, KNN with machine learned distances is also widely used in the field of time series classification [42]. In the field of time series classification, general distance functions, such as euclidean distance, Hamming distance and other distance functions calculated similarity by corresponding position elements. They are not suitable for time series classification. The reasons for this are: when the same phenomenon is observed many times, we can't expect it to always occur at the same time and location. And the duration of the event may be slightly different [43]. However, there are still some distance functions to solve this problem. For example, Sakoe and Chiba proposed a dynamic time warping (DTW) to calculate the distance between spoken recognition time series [44]. Specifically, it first obtains the general principle of time normalization by using the time distortion function. Then it deduces the symmetric and asymmetric forms of distance function from this principle. Finally, it uses slope constraint to limit the slope of warpage function, so as to improve the ability to distinguish between different classes. The nearest neighbor classifier based on DTW and its variants have achieved great success in time series classification because they consider the unique time features in time series data.

## 2.3 Data Collection and Reachable Distance

From the development of distance functions, different real applications often need different distance functions, which have given birth to various distance functions. It is true that these distance functions are ideal and may not output reachable distances. The main reason is that the data miner and data collector are blind to each other. In other words, data miners believe that the training data are satisfied to their data mining applications. And data collectors take data as detailed as possible, so as to support much more data mining applications. In this way, some natural separation information can be merged into databases, see Case I in the introduction section.

From extant data mining applications, both data collectors and data miners are unaware of that there may be an unbridgeable gap between two data points, i.e., the euclidean distance between two data points is not the reachable distance between in them. This must lead to that the performance is decreased.

Different from current distance functions, this research proposes a reachable distance function, aiming at that the intraclass data points are always closer than those interclass data points in training datasets. The reachable distance function finds a clue to developing more suitable distance functions.

## 3 APPROACH

In this article, we use lowercase letters, lowercase bold letters, and uppercase bold letters to represent scalars, vectors, and matrices, respectively. Assume a given sample data set $\mathbf{X} \in \mathbb{R}^{n \times d}$, where $n$ and $d$ represent the number of samples and the number of features, respectively. $a_j$ represents the $j$th element in vector $\mathbf{a}$. And let $\mathbf{c}_a$ be the center point of the class in which sample point $\mathbf{a}$ is located, $\mathbf{c}_b$ be the center point of the class in which sample point $\mathbf{b}$ is located. The symbols used in this paper are shown in Table 2.

TABLE 2
Notations Used in This Paper

| Notations | Descriptions |
|---|---|
| $\mathbf{X} \in \mathbb{R}^{n \times d}$ | Training set with $n$ samples and $d$ features. |
| $\mathbf{X}_{test} \in \mathbb{R}^{m \times d}$ | Test set with $m$ samples and $d$ features. |
| $\mathbf{X}_{label} \in \mathbb{R}^{1 \times n}$ | Class label of training data |
| $\mathcal{Y}_S$ | Class label of test data. |
| $\mathbf{a}$ | Sample $\mathbf{a}$. |
| $\mathbf{b}$ | Sample $\mathbf{b}$. |
| $\mathbf{a}_j$ | The $j$th element in vector $\mathbf{a}$. |
| $\mathbf{c}_a$ | The center point of the class in which sample point $\mathbf{a}$ is located. |
| $\mathbf{c}_b$ | The center point of the class in which sample point $\mathbf{b}$ is located. |
| $\mathbf{c}_c$ | The center point of the class in which sample point $\mathbf{e}$ is located. |
| $\mathbf{c}_1, \mathbf{c}_2, \mathbf{c}_3, \ldots, \mathbf{c}_c$ | The center point of the first to $c$th classes. |
| $d(\mathbf{a}, \mathbf{b})$ | Euclidean distance between $\mathbf{a}$ and $\mathbf{b}$. |
| $mean()$ | Mean function. |
| $G_j$ | the $j$th class data in the training set. |
| $N(Z_0)$ and $N(Z)$ | K-nearest neighbor set under $Z_0$ distance and $Z$ distance. |
| $\mu$ and K | Adjustable parameters. |

## 3.1 Reachable Distance Function

In the field of data mining, distance functions are often used to measure the affinity relation of data points, such as classification and clustering. In the KNN classification, the euclidean distance is most commonly-used to calculate the distance between two points to obtain a neighbor. It could be true that the quality of the KNN classification is largely dependent on the distance formula. If the distance metric formula measures the distance from data of same class is far away, this will result in misclassification. When we look at the traditional distance function, we find that the distance metric only involves the information that the sample point itself has (the value of the feature), and there are many other pieces of information that are not considered. For example, in the classification, each sample has its own classification information except its own feature information.

Considering the above problem, we want to lead some information about the class (e.g., the class center point) into the classification. When we find the class center point for each class, we can use some new distance formulas for classification. We can first get the nearest center point classification (NCP), as shown below

$$d_{min} = \min\{d(\mathbf{t}, \mathbf{c}_1), d(\mathbf{t}, \mathbf{c}_2), \ldots, d(\mathbf{t}, \mathbf{c}_c)\}, \quad (10)$$

where $d(\mathbf{t}, \mathbf{c}_1)$ represents the euclidean distance from $\mathbf{t}$ to $\mathbf{c}_1$, $\mathbf{t}$ represents the test data point, and $\mathbf{c}_1, \mathbf{c}_2, \mathbf{c}_3, \ldots, \mathbf{c}_c$ represents the center point of the first to $c$th classes. From Eq. (10), it can be taken as that in the process of classification, we do not need to set any parameters like K-nearest neighbor classification (such as the selection of K value). In practical applications, it is only necessary to request the distance of the test data to the center point of each class. Then, which distance is closest, the class in which the center point is located is predicted as the class label of the test data.

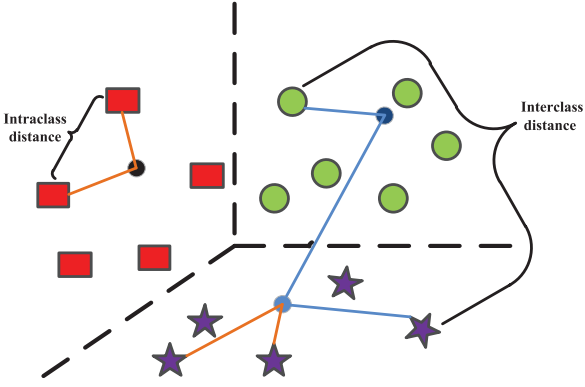

Fig. 3. The schematic diagram of Eq. (11).

Although the above-described distance function (i.e., Eq. (10)) takes the characteristics of the class into account, it is still based on the euclidean distance to some extent. In addition, the method is poorly separable for the calculated distance, because different class centers may be close to each other or different class centers are the same distance from the test data points, and the classification effect of the algorithm will not be good. In summary, in this paper, we propose a reachable distance function for KNN classification as follows.

**Definition 1.** *Let* **a** *and* **b** *be two sample points,* $\mathbf{c}_a$ *the center point of the class in which sample point* **a** *is located, and* $\mathbf{c}_b$ *the center point of the class in which sample point* **b** *is located. The* $Z_0$ *distance between* **a** *and* **b** *is defined as*

$$Z_0(\mathbf{a}, \mathbf{b}) = d(\mathbf{a}, \mathbf{c}_a) + d(\mathbf{b}, \mathbf{c}_b) + \mu * d(\mathbf{c}_a, \mathbf{c}_b), \quad (11)$$

*where* $d()$ *is the euclidean distance between two points.*

It can be seen from Eq. (11) that if **a** and **b** do not belong to the same class, then their distance is farther through the $Z_0$ function. In other words, by the calculation of the Eq. (11), it can make the distance between points of same class smaller than the distance between sample points of different class. In this way, the separability of the class is greatly increased. It is undeniable that this distance measurement function has changed our previous perception of distance. It should be noted that method makes the classes more separable, but it is true that it increases the distance between similar sample points compared to the traditional method. Because it introduces a class center point, it can be proved from Fig. 3, i.e., it makes the original straight line distance into a polyline distance. In response to this small defect, we can improve the $Z_0$ distance in Eq. (11) as follows.

**Definition 2.** *The* $Z$ *distance (reachable distance) function is defined as*

$$Z(\mathbf{a}, \mathbf{b}) = d(\mathbf{a}, \mathbf{b}) + \mu * d(\mathbf{c}_a, \mathbf{c}_b), \quad (12)$$

*where* $\mu$ *is a parameter.*

It can be seen from Eq. (12) that the distance function can also make the distance between sample points of same class smaller than the distance between sample points of different class, which is established on a suitable $\mu$ value. This method not only inherits the advantages of Eq. (11), but also compensates for its shortcomings to some extent, i.e., the distance between two points in the same class is closer. In other word, it not only makes the data points of different classes farther, but also makes the data points of the same class closer.

---

**Algorithm 1.** Pseudo Code for NCP-KNN

---

**Input:** Training set $\mathbf{X} \in \mathbb{R}^{n \times d}$, Labels of the training data set $\mathbf{X}_{label} \in \mathbb{R}^{1 \times n}$, Test Data $\mathbf{X}_{test} \in \mathbb{R}^{m \times d}$ and K;
**Output:** Class label $\mathcal{Y}_S$ of test data;
1: **for** $i = 1 \rightarrow m$ **do**
2:     **for** $j = 1 \rightarrow c$ **do**
3:        $\{\mathbf{c}_1, \mathbf{c}_2, \mathbf{c}_3, \ldots, \mathbf{c}_c\}$ is calculated by $mean(G_j)$;
4:        $d(\mathbf{X}_{test}^i, \mathbf{c}_j) = \sqrt{(\mathbf{X}_{test}^i - \mathbf{c}_j)^T (\mathbf{X}_{test}^i - \mathbf{c}_j)}$;
5:     **end**
6:     Calculate $d_{min} = \min(d(\mathbf{X}_{test}^i, \mathbf{c}_1), \ldots, d(\mathbf{X}_{test}^i, \mathbf{c}_c))$;
7:     We can get the class label $\mathcal{Y}_S$ of the test data according to the class label corresponding to $d_{min}$;
8: **end**

---

---

**Algorithm 2.** Pseudo Code for $Z_0$-KNN and Z-KNN

---

**Input:** Training set $\mathbf{X} \in \mathbb{R}^{n \times d}$, Labels of the training data set $\mathbf{X}_{label} \in \mathbb{R}^{1 \times n}$, Test Data $\mathbf{X}_{test} \in \mathbb{R}^{m \times d}$, K and $T$;
**Output:** Class label $\mathcal{Y}_S$ of test data;
1: **for** $i = 1 \rightarrow m$ **do**
2:     **for** $j = 1 \rightarrow c$ **do**
3:        $\{\mathbf{c}_1, \mathbf{c}_2, \mathbf{c}_3, \ldots, \mathbf{c}_c\}$ is calculated by $mean(G_j)$;
4:     **end**
5:     Suppose the class center point of $\mathbf{X}_{test}^i$ is $mean(\mathbf{X})$ ;
6:     **for** $l = 1 \rightarrow n$ **do**
7:        Calculate $Z_0(\mathbf{X}_{test}^i, \mathbf{X}^l)$ and $Z(\mathbf{X}_{test}^i, \mathbf{X}^l)$ by Eqs. (11) and (12);
8:     **end**
9:     **if** $T == Z_0$-KNN **then**
10:        $N(Z_0) = \min_{1 \rightarrow K}(Z_0(\mathbf{X}_{test}^i, \mathbf{X}^1), \ldots, Z_0(\mathbf{X}_{test}^i, \mathbf{X}^n))$;
11:     **end**
12:     **if** $T == Z$-KNN **then**
13:        $N(Z) = \min_{1 \rightarrow K}(Z(\mathbf{X}_{test}^i, \mathbf{X}^1), \ldots, Z(\mathbf{X}_{test}^i, \mathbf{X}^n))$;
14:     **end**
15:     We get the class label $\mathcal{Y}_S$ of the test data according to majority rule on $N(Z_0)$ or $N(Z)$;
16: **end**

---

We wrote the pseudocode of NCP-KNN based on Eq. (10) in Algorithm 1, and the pseudocode of $Z_0$-KNN and Z-KNN based on Eqs. (11) and (12) in Algorithm 2. It should be noted that when a new test data is given, we do not know its class center. Therefore, in step 5 of Algorithm 2, we assume that its class center is the center of the overall data. To some extent, it may lead to a bias to predict the test data as the class closest to the overall data center. When the distance between the centers of all classes in the data and the center of the overall data is almost equal, this kind of bias will hardly exist, because each class center is treated equally. When the centers of all classes in the data are not equidistant from the center of the overall data, it may lead to a more biased prediction of the class closest to the overall data center. The reason is: for the class center closest to the overall data center, all data points in this class are not necessarily close to the overall data center, i.e., when calculating the K nearest neighbors of the test data according to the distance function, there are still some nearest neighbors in other classes to be selected. At this time, the selection of K value and classification rules (such as fixed K value and majority) are very important. Of course, when the data is class unbalanced or the distance between the centers of all classes in the data and the center of the overall data is very unequal (the distance difference is very great). At this time, it belongs to an extreme case, the K nearest neighbors of the test sample only come from the data in the class closest to the overall data center, unless the K value is very large, i.e., K is greater than the number of data in the class closest to the overall data center.

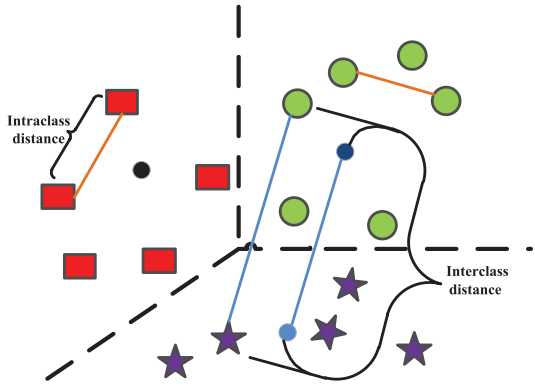

Fig. 4. The schematic diagram of Eq. (12).

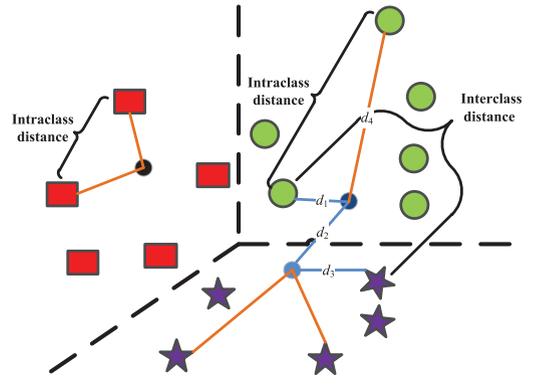

(a) Case of Eq. (11).

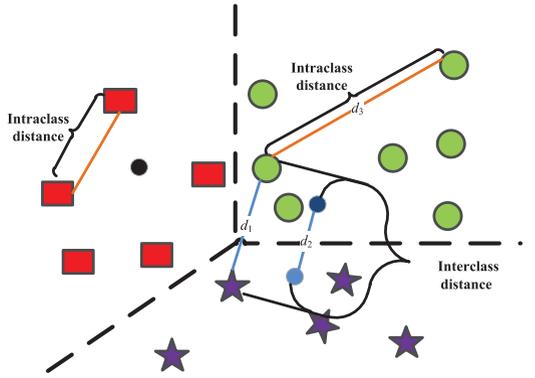

(b) Case of Eq. (12).

Fig. 5. The schematic diagram of the problem with the new distance function.

To sum up, the proposed algorithm is more suitable for the case where the distance difference between all class centers and the overall data center is not particularly large. Its performance will be affected by the distance between each class center and the overall data center, because the distance between each class center and the overall data center will affect the search of K nearest neighbors of the test data. In addition, its performance is also affected by the K value and $\mu$ value, because the K value and $\mu$ value will affect the search for the K nearest neighbors of the test data.

In order to better understand the proposed reachable distance, we describe it with Figs. 3, 4, and 5. We focus on the characteristics of the two new distance functions according to Figs. 3 and 4. Fig. 3 shows the distance between data points in Eq. (11). From Fig. 3, we can find that the distance between two data points of the same class is not a straight line distance, it needs to pass the class center point. The shape of the distance between two data points in different classes is like a "Z". Fig. 4 shows the distance between data points in Eq. (12). From Fig. 4, we can find that the distance between two data points is the same as the traditional euclidean distance in the same class. In different classes, the distance between two data points includes not only the euclidean distance between them, but also the distance between the center of the class. Fig. 5 shows a situation in practice. For example, in the case of Eq. (11), $d_1 + d_2 + d_3$ is likely to be smaller than $d_1 + d_4$. In the case of Eq. (12), $d_1 + d_2$ is likely to be smaller than $d_3$. The distance between data points of different class may be smaller than the distance between data points of same class. To avoid it, we introduced the parameter $\mu$. The size of the $\mu$ value will affect the measurement of the natural distance between two data points. Usually, when $\mu > 1$, intraclass distance is less than interclass distance. In addition, we can see that the distance between data points of same class in Eq. (11) is greater than the distance between data points of same class in Eq. (12). This is the main difference between Eqs. (11) and (12).

## 3.2 Properties of Z Distance

The Z (reachable) distance function has three basic properties as follows.

**Property 1.** *Nonnegativity:* $z(\mathbf{a}, \mathbf{b}) \geq 0$

**Property 2.** *Symmetry:* $z(\mathbf{a}, \mathbf{b}) = z(\mathbf{b}, \mathbf{a})$

**Property 3.** *Directness:* $z(\mathbf{a}, \mathbf{e}) \leq z(\mathbf{a}, \mathbf{b}) + z(\mathbf{b}, \mathbf{e})$

Now let's prove these three properties.

**Proof.** For nonnegativity, because Z distance is based on euclidean distance, it is obvious that the proposed Z distance (Eqs. (11) and (12)) is consistent. In Eq. (12), only if $\mathbf{a} = \mathbf{b}$, $z(\mathbf{a}, \mathbf{b}) = 0$. □

**Proof.** For symmetry, both Eqs. (11) and (12) are also satisfied, as shown below

$$z(\mathbf{a}, \mathbf{b}) = d(\mathbf{a}, \mathbf{c}_a) + d(\mathbf{b}, \mathbf{c}_b) + d(\mathbf{c}_a, \mathbf{c}_b)$$
$$= z(\mathbf{b}, \mathbf{a}) = d(\mathbf{b}, \mathbf{c}_b) + d(\mathbf{a}, \mathbf{c}_a) + d(\mathbf{c}_b, \mathbf{c}_a)$$
$$= \left[ \sum_{j=1}^{d} (a_j - c_{aj})^2 \right]^{\frac{1}{2}} + \left[ \sum_{j=1}^{d} (b_j - c_{bj})^2 \right]^{\frac{1}{2}}$$
$$+ \left[ \sum_{j=1}^{d} (c_{aj} - c_{bj})^2 \right]^{\frac{1}{2}} \tag{13}$$

$$z(\mathbf{a}, \mathbf{b}) = d(\mathbf{a}, \mathbf{b}) + \mu * d(\mathbf{c}_a, \mathbf{c}_b)$$
$$= z(\mathbf{b}, \mathbf{a}) = d(\mathbf{b}, \mathbf{a}) + \mu * d(\mathbf{c}_b, \mathbf{c}_a)$$
$$= \left[ \sum_{j=1}^{d} (a_j - b_j)^2 \right]^{\frac{1}{2}} + \mu * \left[ \sum_{j=1}^{d} (c_{aj} - c_{bj})^2 \right]^{\frac{1}{2}}. \quad (14)$$

From Eqs. (13) and (14), we can see that the proposed Z-distance has symmetry. □

**Proof.** For Property 3, i.e., the proposed Z distance satisfies the directness in the following two cases.

(1) When data points $\mathbf{a}$, $\mathbf{b}$ and $\mathbf{e}$ belong to the same class. According to Eq. (11), we can get the following formula:

$$z(\mathbf{a}, \mathbf{b}) + z(\mathbf{b}, \mathbf{e}) - z(\mathbf{a}, \mathbf{e})$$
$$= d(\mathbf{a}, \mathbf{c}_a) + d(\mathbf{b}, \mathbf{c}_a) + d(\mathbf{b}, \mathbf{c}_a)$$
$$+ d(\mathbf{e}, \mathbf{c}_a) - d(\mathbf{a}, \mathbf{c}_a) - d(\mathbf{e}, \mathbf{c}_a)$$
$$= 2 * d(\mathbf{b}, \mathbf{c}_a) = 2 * \left[ \sum_{j=1}^{d} (b_j - c_{aj})^2 \right]^{\frac{1}{2}} \geq 0. \quad (15)$$

According to Eq. (12) and trigonometric inequality, we can get the following formula:

$$z(\mathbf{a}, \mathbf{b}) + z(\mathbf{b}, \mathbf{e}) - z(\mathbf{a}, \mathbf{e})$$
$$= d(\mathbf{a}, \mathbf{b}) + d(\mathbf{b}, \mathbf{e}) - d(\mathbf{a}, \mathbf{e})$$
$$= \left[ \sum_{j=1}^{d} (a_j - b_j)^2 \right]^{\frac{1}{2}} + \left[ \sum_{j=1}^{d} (b_j - e_j)^2 \right]^{\frac{1}{2}}$$
$$- \left[ \sum_{j=1}^{d} (a_j - e_j)^2 \right]^{\frac{1}{2}} \geq 0. \quad (16)$$

From Eqs. (15) and (16), we can get that when $\mathbf{a}$, $\mathbf{b}$ and $\mathbf{e}$ belong to the same class, the proposed Z distance (Eqs. (11) and (12)) have the property of directness.

(2) When data points $\mathbf{a}$, $\mathbf{b}$ and $\mathbf{e}$ belong to different classes. According to Eq. (11) and trigonometric inequality, we can get the following formula:

$$z(\mathbf{a}, \mathbf{b}) + z(\mathbf{b}, \mathbf{e}) - z(\mathbf{a}, \mathbf{e})$$
$$= d(\mathbf{a}, \mathbf{c}_a) + d(\mathbf{b}, \mathbf{c}_b) + \mu * d(\mathbf{c}_a, \mathbf{c}_b)$$
$$+ d(\mathbf{b}, \mathbf{c}_b) + d(\mathbf{e}, \mathbf{c}_e) + \mu * d(\mathbf{c}_b, \mathbf{c}_e)$$
$$- d(\mathbf{a}, \mathbf{c}_a) - d(\mathbf{e}, \mathbf{c}_e) - \mu * d(\mathbf{c}_a, \mathbf{c}_e)$$
$$= 2 * d(\mathbf{b}, \mathbf{c}_b) + \mu * d(\mathbf{c}_a, \mathbf{c}_b)$$
$$+ \mu * d(\mathbf{c}_b, \mathbf{c}_e) - \mu * d(\mathbf{c}_a, \mathbf{c}_e)$$
$$= 2 * \left[ \sum_{j=1}^{d} (b_j - c_{bj})^2 \right]^{\frac{1}{2}} + \mu * \left[ \sum_{j=1}^{d} (c_{aj} - c_{bj})^2 \right]^{\frac{1}{2}}$$
$$+ \mu * \left[ \sum_{j=1}^{d} (c_{bj} - c_{ej})^2 \right]^{\frac{1}{2}} + \mu * \left[ \sum_{j=1}^{d} (c_{aj} - c_{ej})^2 \right]^{\frac{1}{2}} \geq 0. \quad (17)$$

According to Eq. (12), we can get the following formula:

$$z(\mathbf{a}, \mathbf{b}) + z(\mathbf{b}, \mathbf{e}) - z(\mathbf{a}, \mathbf{e})$$
$$= d(\mathbf{a}, \mathbf{b}) + \mu * d(\mathbf{c}_a, \mathbf{c}_b)$$
$$+ d(\mathbf{b}, \mathbf{e}) + \mu * d(\mathbf{c}_b, \mathbf{c}_e)$$
$$- d(\mathbf{a}, \mathbf{e}) - \mu * d(\mathbf{c}_a, \mathbf{c}_e)$$
$$= \left[ \sum_{j=1}^{d} (a_j - b_j)^2 \right]^{\frac{1}{2}} + \mu * \left[ \sum_{j=1}^{d} (c_{aj} - c_{bj})^2 \right]^{\frac{1}{2}}$$
$$+ \left[ \sum_{j=1}^{d} (b_j - e_j)^2 \right]^{\frac{1}{2}} + \mu * \left[ \sum_{j=1}^{d} (c_{bj} - c_{ej})^2 \right]^{\frac{1}{2}}$$
$$- \left[ \sum_{j=1}^{d} (a_j - e_j)^2 \right]^{\frac{1}{2}} - \mu * \left[ \sum_{j=1}^{d} (c_{aj} - c_{ej})^2 \right]^{\frac{1}{2}} \geq 0. \quad (18)$$

According to Eqs. (17) and (18), when $\mathbf{a}$, $\mathbf{b}$ and $\mathbf{e}$ belong to different classes, the Z distance (Eqs. (11) and (12)) has the property of directness. In conclusion, the proposed Z distance satisfies the property of directness.□

**Corollary 1.** *Intraclass distance is less than interclass distance.*

**Proof.** For Eq. (11), if data points $\mathbf{a}$ and $\mathbf{b}$ belong to the same class, then the Z distance between them is the following formula:

$$z(\mathbf{a}, \mathbf{b}) = d(\mathbf{a}, \mathbf{c}_a) + d(\mathbf{b}, \mathbf{c}_a), \quad (19)$$

where $c_a$ is the class center of data points $\mathbf{a}$ and $\mathbf{b}$. If data points $\mathbf{a}$ and $\mathbf{e}$ belong to different classes, the Z distance between them is the following formula:

$$z(\mathbf{a}, \mathbf{e}) = d(\mathbf{a}, \mathbf{c}_a) + d(\mathbf{e}, \mathbf{c}_e) + \mu * d(\mathbf{c}_a, \mathbf{c}_e). \quad (20)$$
□

From Eqs. (19) and (20), we can see that as long as the following equations are proved to be true:

$$d(\mathbf{e}, \mathbf{c}_e) + \mu * d(\mathbf{c}_a, \mathbf{c}_e) > d(\mathbf{b}, \mathbf{c}_a). \quad (21)$$

Obviously, the interclass distance is one more natural distance than the intraclass distance, i.e., $\mu * d(\mathbf{c}_a, \mathbf{c}_e)$. If the value of parameter $\mu$ is infinite, then Eq. (21) is sure to hold. When $\mu$ takes a very small value, Eq. (21) may not hold. Therefore, if the value of parameter $\mu$ is large, then Eq. (11) satisfies the characteristic that intraclass distance is less than interclass distance.

Similarly, for Eq. (12), we only need to prove that the following formula holds:

$$d(\mathbf{a}, \mathbf{e}) + \mu * d(\mathbf{c}_a, \mathbf{c}_e) > d(\mathbf{a}, \mathbf{b}). \quad (22)$$

We can see that, as in Eq. (11), if the value of parameter $\mu$ is large, Eq. (12) satisfies the characteristic that intraclass distance is less than interclass distance.

### 3.3 Comparative Analysis

The Z distance is based on euclidean distance, which can be regarded as an improvement of euclidean distance. Compared with euclidean distance, Z distance not only considers the natural distance, but also makes the distance between different classes greater than the distance between data in the same class. The properties and functions of euclidean distance and Z distance are listed in Tables 3 and 4, respectively. In addition, For the traditional KNN algorithm based on euclidean distance, because each test data point needs to calculate the distance with all training data, its time complexity is $O(n * d)$, where $n$ represents the number of training samples and $d$ represents the dimension of data. If there are $m$ test data, its time complexity is $O(m * n * d)$. Similarly, the difference between Z-KNN and KNN method lies in the distance function (i.e., Z distance and euclidean distance). The Z distance is based on euclidean distance, and the calculation times of Z-KNN and KNN in finding k nearest neighbors are the same. Therefore, the time complexity of Z-KNN is still $O(m * n * d)$.

TABLE 3
Properties of Euclidean Distance and Z Distance

| Distance function | Nonnegativity | Symmetry | Directness | Intraclass Distance is Less Than Interclass Distance |
|---|---|---|---|---|
| Euclidean distance | ✓ | ✓ | ✓ | × |
| Z distance | ✓ | ✓ | ✓ | ✓ |

In Fig. 6, we show the comparison between the proposed Z distance (i.e., Eqs. (11) and (12)) and euclidean distance. From Fig. 6, we can see that using euclidean distance to calculate the distance between original data does not make classes separable, i.e., intraclass distance may be larger than interclass distance. The proposed Z distance (Eq. (11)) can increase the interclass distance, which leads to higher separability of different classes. However, it also increases the intraclass distance, making similar samples more dispersed. The Z distance (Eq. (12)) not only increases the interclass distance, but also makes the intraclass distance constant (the same as the euclidean distance). Therefore, Z distance (Eq. (12)) has the best class separability.

## 4 EXPERIMENTS

In order to verify the validity of the new distance functions, we compare the KNN classification accuracy of the new distance functions and the 6 other comparison algorithms with 12 data sets[1,2] (as shown in Table 5).

### 4.1 Experiment Settings

We download the data sets for our experiments from the datasets website, which includes 4 binary datasets and 8 multiclassification datasets. We divide each dataset into a training set and a test set by ten-fold cross-validation (i.e., we divide the data set into 10 parts, 9 of which are used as training sets, and the remaining one is used as a test set, which is sequentially cycled until all data have been tested). The comparison algorithm is introduced as follows during the experiment:

KNN [45]: It's the traditional KNN algorithm, and we don't have to do anything during the training phase. In the test phase, for each test data point, we find its K neighbors in the training data according to the euclidean distance. Then, the class label with the highest frequency of class in the K neighbors is selected as the final class label of the test data.

Nearest class center point for KNN (NCP-KNN): This method is the most basic algorithm after introducing the class center point. In the training phase, a class center point is obtained for the training data in each class. In the test phase, we calculate the distance between each test data and the center point of the training process. The class of the nearest class center point is the class label of the test data.

Coarse to fine K nearest neighbor classifier (CFKNN) [46]: In this method, a new metric function is proposed, which is expressed linearly by test data. Specifically, it first uses training data to represent test data through least squares loss. Then it gets the relational metric matrix by solving the least squares loss. Finally, it uses the new metric matrix to construct a new distance function. It classifies the test data according to the new distance function and major rule.

Local mean representation-based k-nearest neighbor classifier (LMRKNN) [47]: It is an improved KNN method based on local mean vector representation. Specifically, it first finds K neighbors in each class and constructs a local mean vector. Then it uses these local mean vectors to represent each test data and obtains a relationship measurement matrix. Finally, it uses the matrix to construct a new distance function for KNN.

Graph regularized k-local hyperplane distance nearest neighbor algorithm (GHKNN) [48]: This method is a local hyperplane nearest neighbor algorithm based on multi-kernel learning. Specifically, it first constructs six sequence based feature descriptors, and then learns the weight of features. Finally, graph regularized k-local hyperplane KNN is used to classify the subcellular localization of noncoding RNA.

Minkowski distance based fuzzy k nearest neighbor algorithm (MDFKNN) [49]: This method uses Minkowski distance to replace euclidean distance, which avoids the invalidity of euclidean distance to high-dimensional data. Specifically, it first uses Minkowski distance to calculate the nearest neighbor data similar to each test data. Then it applies fuzzy weight to the nearest neighbor data. Finally, it uses weighted average to achieve more accurate prediction of data.

A Weighted Mutual k-Nearest Neighbour (WKNN) [50]: This method can eliminate the influence of noise and pseudo neighbors in data. Specifically, it uses mutual domain and distance weighted voting to weaken the influence of distant neighbors. In addition, after removing outliers, the dataset will be refined, which makes the algorithm more inclined to consider those nearest neighbors.

TABLE 4
Function Comparison Between Euclidean
Distance and Z Distance

| Distance Function | Measuring Affinity | Reachable Distance |
|---|---|---|
| Euclidean distance | ✓ | × |
| Z distance | ✓ | ✓ |

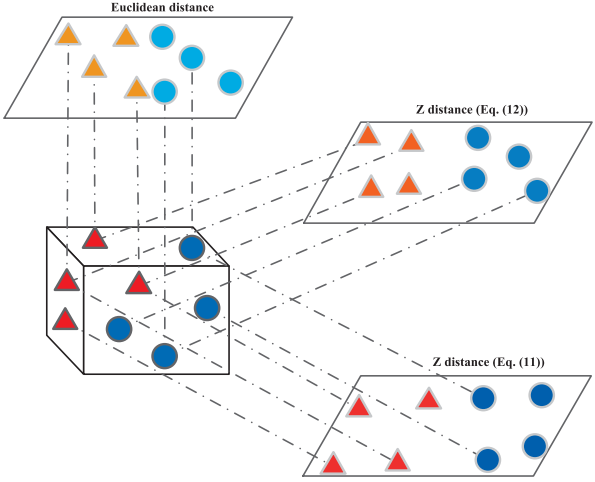

Fig. 6. The schematic diagram of three distance function comparisons.

TABLE 5
The Information of the Data Sets

| Datasets | Number of Samples | Dimensions | Classes |
|---|---|---|---|
| Banknote | 1372 | 4 | 2 |
| Cnae | 1080 | 856 | 9 |
| Drift | 1244 | 129 | 5 |
| Secom | 1567 | 590 | 2 |
| Ionosphere | 351 | 34 | 2 |
| Usps | 9298 | 256 | 10 |
| Yeast | 1484 | 1470 | 10 |
| Letter | 20000 | 16 | 26 |
| Movements | 360 | 90 | 15 |
| Multiple | 2000 | 649 | 10 |
| Statlog | 6435 | 36 | 6 |
| German | 1000 | 20 | 2 |

$Z_0$-KNN: It is the traditional KNN method based on the $Z_0$ distance function (i.e., Eq. (11)). During the training process, we calculate the center point of each class in the training data. In the test process, we find K neighbors from training data according to Eq. (11). And then, we use the majority rule to predict the class label of test data.

Z-KNN: It is the traditional KNN method based on the Z distance function (i.e., Eq. (12)). It is basically the same as $Z_0$-KNN's training process and testing process. The only difference is that it is based on Eq. (12).

For the above algorithms, we did a series of experiments. Specifically, for each dataset, we test all the algorithms by setting different K values (i.e., 1-10), where the NCP- KNN algorithm has no K parameter, so we have performed 10 experiments for it. It is convenient for us to put all the algorithms in one subgraph. Finally, we measure their performance based on classification accuracy. In addition, in the case of K = 5, we performed 10 experiments on all algorithms to preserve the average classification accuracy and standard deviation. Finally, for the binary classification dataset, we not only calculated their classification accuracy, but also calculated their Sensitivity (Sen) and Specificity (Spe).

The accuracy(Acc) and standard deviation(std) are calculated by the following equations respectively:

$$Acc = X_{correct}/X_{total}, \qquad (23)$$

where $X_{correct}$ represents the number of test data that is correctly classified, and $X_{total}$ represents the total number of test data.

$$std = \sqrt{\frac{1}{n}\sum_{i=1}^{n}\left(Acc_i - \widetilde{Acc}\right)^2}, \qquad (24)$$

where $n$ represents the number of experiments, $Acc_i$ represents the classification accuracy of the $i$th experiment, and $\widetilde{Acc}$ represents the average classification accuracy of the experiment. The smaller the $std$, the more stable the algorithm is.

### 4.2 Binary Classification

Table 6 shows the classification performance of all algorithms on the binary datasets. We can get the some result, i.e., the Z-KNN algorithm achieves the best results, and Ncp-KNN performs the worst. Specifically, on the German dataset, the classification accuracy of the Z-KNN algorithm is 7.14% higher than the traditional KNN algorithm. Because

TABLE 6
Experimental Results for Binary Data Sets

| Datasets | Banknote | | | German | | | Ionosphere | | | Secom | | |
|---|---|---|---|---|---|---|---|---|---|---|---|---|
| | ACC | SEN | SPE | ACC | SEN | SPE | ACC | SEN | SPE | ACC | SEN | SPE |
| KNN | 54.59 | 49.02 | 59.06 | 61.56 | 80.29 | 18.50 | 58.20 | 77.78 | 23.97 | 92.26 | 0.03 | 98.77 |
| NCP-KNN | 52.15 | 47.87 | 57.35 | 56.29 | 67.29 | 31.00 | 52.91 | 56.53 | 46.43 | 64.86 | 56.35 | 42.95 |
| CFKNN | 52.53 | 48.12 | 53.45 | 61.23 | 77.64 | 22.93 | 56.92 | 66.23 | 34.64 | 90.36 | 4.04 | 95.95 |
| LMRKNN | 55.27 | 49.65 | 55.45 | 56.67 | 66.33 | 34.13 | 52.81 | 56.63 | 46.04 | 75.37 | 25.67 | 76.34 |
| GHKNN | 55.17 | 49.67 | 59.58 | 55.64 | 64.99 | 33.83 | 58.26 | 78.71 | 21.75 | 92.22 | 2.79 | 98.58 |
| MDFKNN | 55.11 | 39.84 | 57.44 | 59.06 | 71.97 | 28.93 | 56.81 | 71.78 | 30.08 | 82.85 | 14.72 | 86.05 |
| WKNN | 55.35 | 49.87 | 59.74 | 60.27 | 76.03 | 23.43 | 57.29 | 75.64 | 24.52 | 88.15 | 7.98 | 93.85 |
| $Z_0$-KNN | 54.45 | 48.52 | 57.09 | 56.42 | 67.57 | 30.00 | 58.40 | 79.56 | 25.79 | 75.94 | 22.69 | 75.35 |
| Z-KNN | 56.56 | 51.15 | 60.89 | 68.70 | 93.00 | 0.06 | 60.68 | 79.87 | 21.43 | 92.98 | 0.01 | 99.25 |

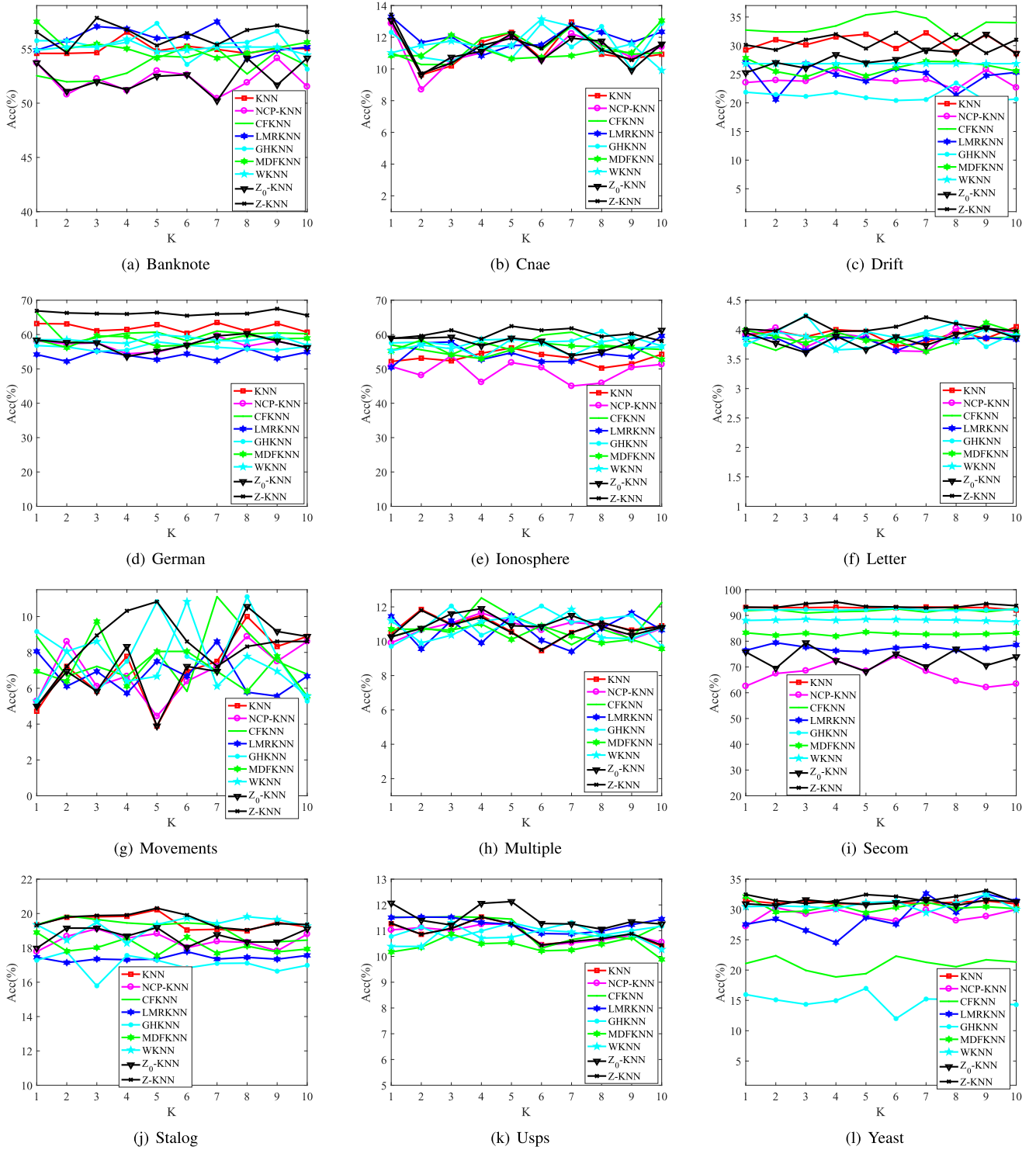

Fig. 7. The classification accuracy on the 12 datasets with different K values.

the euclidean distance used in the traditional KNN does not take into account the natural distance, and the separability of classes is not high. CFKNN uses the relationship matrix between test data and training data to construct a new measurement function, which still does not take into account the natural distance between the data. LMRKNN uses the local mean vector in each class to construct a new distance function. Although it takes into account the local structure of the data, it does not take into account the separability between classes. GHKNN improves KNN through the graph regularization term of multi-kernel learning. It considers the global structure information of data, and it is more suitable for noncoding RNA to locate cells. MDFKNN not only uses Minkowski distance to improve KNN, but also uses the method of applying weight to k-nearest neighbors to reduce the importance of distant neighbors like WKNN. To sum up, the above comparison algorithms do not use reachable distance and ignore the class characteristics of data. The proposed Z-KNN not only considers the natural distance in the data, but also makes the intraclass distance larger, which

TABLE 7
Accuracy (Mean $\pm$ Standard Deviation) Statistical Results on Multi-Class Datasets

| Datasets | Cnae | Drift | Usps | Yeast | Letter | Movements | Multiple | Statlog |
|---|---|---|---|---|---|---|---|---|
| KNN | 10.65 ± 0.01 | 28.78 ± 0.02 | 10.43 ± 0.00 | 30.86 ± 0.00 | 3.87 ± 0.01 | 7.26 ± 0.02 | 10.02 ± 0.01 | 19.35 ± 0.00 |
| NCP-KNN | 10.56 ± 0.01 | 25.88 ± 0.01 | 10.53 ± 0.00 | 29.25 ± 0.01 | 3.84 ± 0.00 | 6.39 ± 0.01 | 9.95 ± 0.01 | 18.32 ± 0.00 |
| CFKNN | 11.52 ± 0.01 | 34.49 ± 0.01 | 10.43 ± 0.01 | 21.57 ± 0.01 | 3.82 ± 0.00 | 7.42 ± 0.01 | 10.34 ± 0.01 | 20.37 ± 0.02 |
| LMRKNN | 12.18 ± 0.02 | 26.37 ± 0.02 | 11.61 ± 0.01 | 28.43 ± 0.01 | 3.86 ± 0.01 | 6.97 ± 0.01 | 10.78 ± 0.01 | 17.13 ± 0.01 |
| GHKNN | 11.55 ± 0.01 | 21.26 ± 0.01 | 11.03 ± 0.00 | 14.86 ± 0.01 | 3.89 ± 0.00 | 8.03 ± 0.02 | 10.96 ± 0.01 | 17.03 ± 0.01 |
| MDFKNN | 11.39 ± 0.01 | 27.16 ± 0.01 | 10.51 ± 0.01 | 30.40 ± 0.01 | 3.87 ± 0.00 | 7.03 ± 0.01 | 10.38 ± 0.01 | 17.75 ± 0.01 |
| WKNN | 11.44 ± 0.01 | 26.85 ± 0.01 | 10.73 ± 0.01 | 30.98 ± 0.01 | 3.88 ± 0.01 | 7.42 ± 0.02 | 10.74 ± 0.01 | 19.39 ± 0.01 |
| $Z_0$-KNN | 11.76 ± 0.01 | 24.28 ± 0.01 | 11.52 ± 0.00 | 31.67 ± 0.01 | 3.86 ± 0.01 | 6.94 ± 0.01 | 10.50 ± 0.00 | 18.77 ± 0.00 |
| Z-KNN | 12.59 ± 0.01 | 31.43 ± 0.01 | 11.53 ± 0.00 | 31.40 ± 0.00 | 4.06 ± 0.00 | 9.72 ± 0.02 | 10.80 ± 0.00 | 20.17 ± 0.00 |

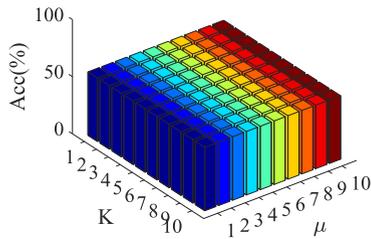

(a) Banknote

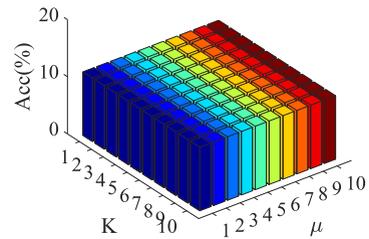

(b) Cnae

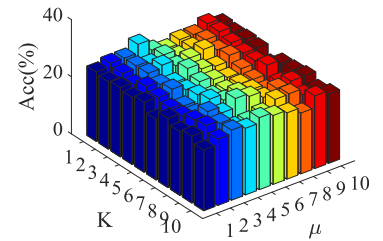

(c) Drift

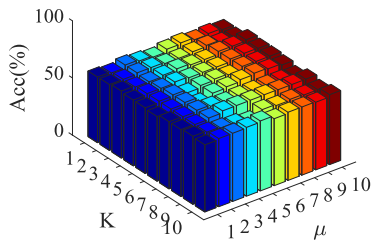

(d) German

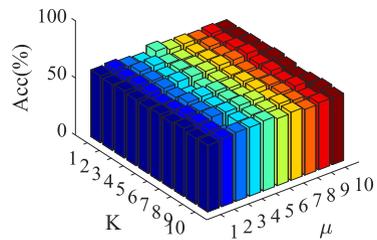

(e) Ionosphere

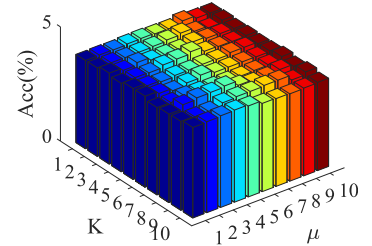

(f) Letter

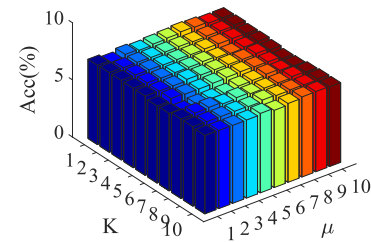

(g) Movements

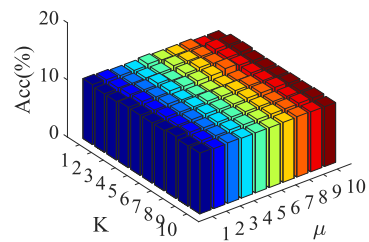

(h) Multiple

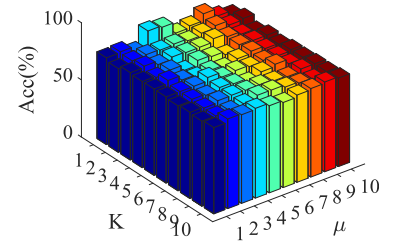

(i) Secom

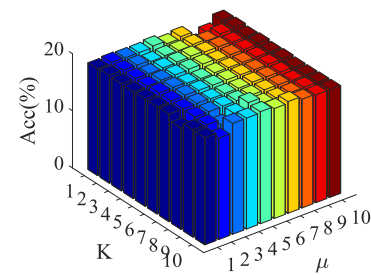

(j) Stalog

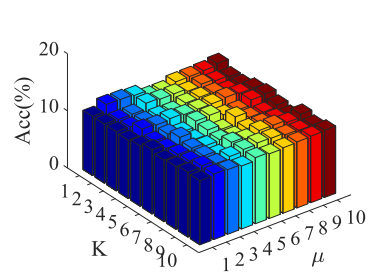

(k) Usps

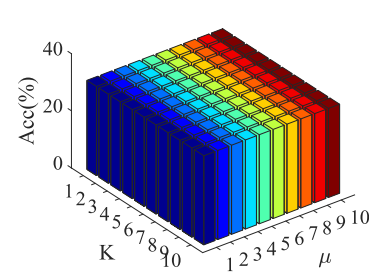

(l) Yeast

Fig. 8. The classification accuracy of different K and $\mu$ parameter (in Eq. (11)) values on the dataset.

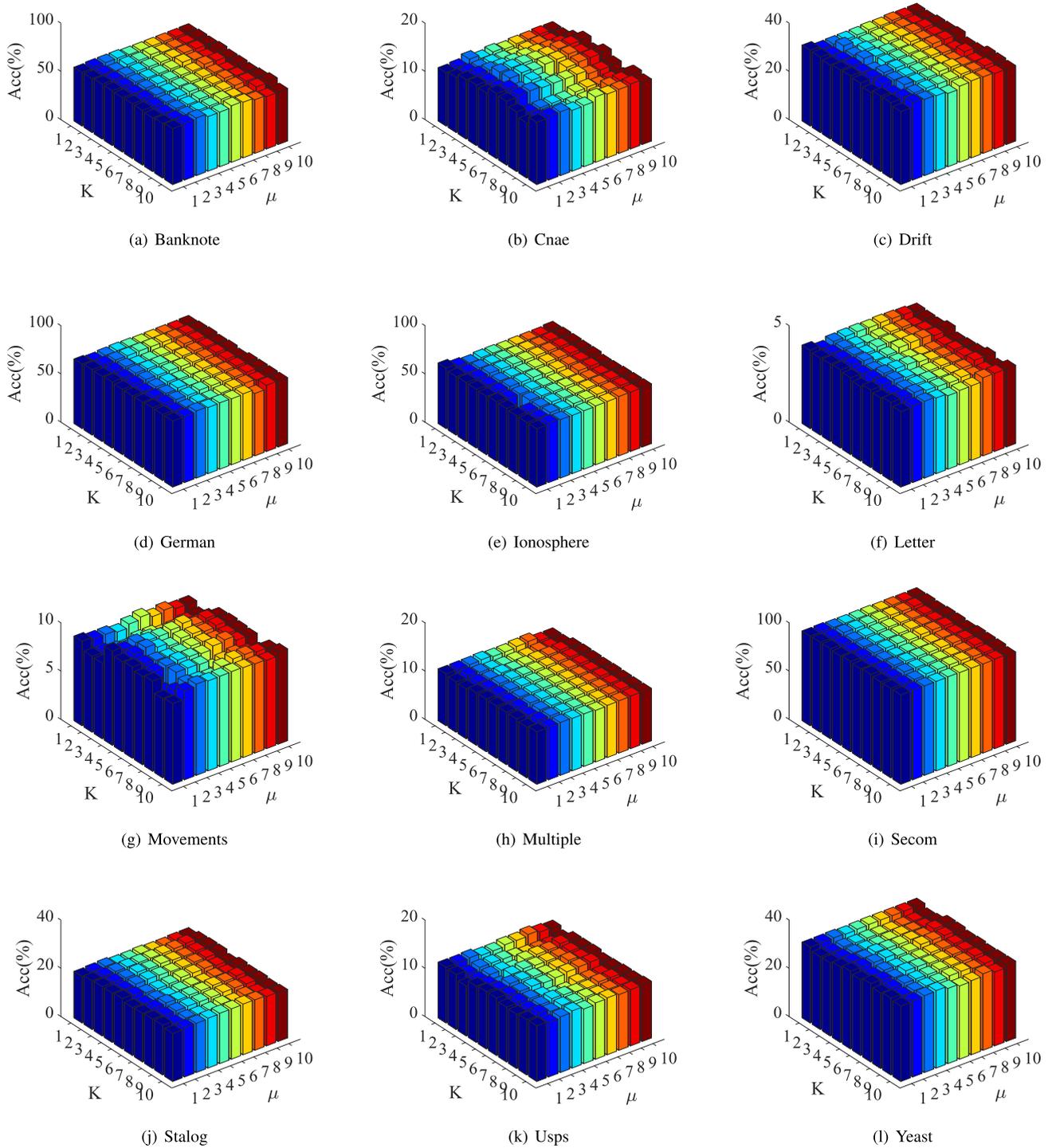

Fig. 9. The classification accuracy of different K and $\mu$ parameter (in Eq. (12)) values on the dataset.

improves the separability of the data and achieves better results.

## 4.3 Multiple Classification

Fig. 7 shows the classification accuracy of all algorithms on 12 data sets as K value. Specifically, we can see that the performance of the Z-KNN algorithm is best in some cases from Fig. 7. The NCP-KNN algorithm has the worst effect, and the overall effect of the $Z_0$-KNN algorithm is not satisfactory, but it achieves the best effect on the Usps dataset, which shows that after we introduce the class feature information, it has a certain effect. The effect of Z-KNN is sufficient to prove that we are looking for a distance function with "high cohesion, low coupling" is very necessary for classification. For the traditional KNN algorithm, its effect is better than the NCP-KNN algorithm, which shows that only considering class information is unreliable. In addition, we also see that Z-KNN does not perform best on some data sets. There are two main reasons for this phenomenon: 1. Different K values will affect the performance of the algorithm.

2. Z-KNN mainly considers the information lost during data collection (such as case 2), which is a reachable distance. The actual dataset may not lose the original information in the collection. At this time, the advantages of Z-KNN can not be shown.

Table 7 shows the average classification accuracy and standard deviation of the algorithm on the multi-class dataset. From Table 7, we can see that the Z-KNN algorithm achieves the best performance on the multi-class dataset except the Yeast dataset. The worst performer is the NCP-KNN algorithm. In addition, we can also see that the $std$ of all algorithms is relatively small, i.e., their stability is very good.

## 4.4 Parameter Sensitivity

In Eqs. (11) and (12), there is a parameter $\mu$, which determines the size of the natural distance. Different $\mu$ value will affect the distance calculation between training data and test data, thus affecting the selection of nearest neighbors. If $\mu$ takes a small value, it may not play the role of measuring natural distance at all, if $\mu$ takes a large value, which may greatly weaken the unnatural distance between samples. Therefore, we set up experiments with different K values and different $\mu$ values. As shown in Figs. 8 and 9, we can see that in most cases, the value of $\mu$ has an impact on the performance of classification. Specifically, on the Drift, Cnae, and Movements data sets, the accuracy rate varies greatly under different $\mu$ values. This shows that one has to adjust the value of parameter $\mu$ carefully. In addition, on some data sets, such as Banknote and Yeast data sets, parameters K and $\mu$ have little impact on the performance of the algorithm. This shows that on the one hand, the selection of K value does not have a great impact on these data sets. On the other hand, this shows that there may be no insurmountable natural distance in these data sets, i.e., there is no information in the data at the time of data collection. Therefore, K value and $\mu$ value have no significant effect on these data sets.

## 5 Conclusion

This paper has proposed a new distance function, reachable distance, or Z distance. Specifically, it takes the class attribute into account in the distance function, and uses the distance between the class center points to measure the natural distance in the data. In addition, it is an reachable distance, and it makes the interclass distance must be greater than the intraclass distance. In the experiment, the KNN based on Z distance (i.e., Z-KNN) exceeds the advanced comparison algorithm in terms of classification accuracy.

In the future work, we plan to proceed from the following three points as follows:

1. Finding one or more better distance functions to make the K-nearest neighbor classification algorithm achieve better performance.
2. Applying this idea to other classification algorithms to find distance functions that are suitable for other classification algorithms.
3. We will find a new distance function to apply to clustering, it is very challenging and interesting.

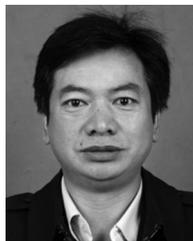

**Shichao Zhang** (Senior Member, IEEE) received the PhD degree from Deakin University, Australia. He is a China National Distinguished professor with the Central South University, China. His research interests include data mining and Big Data. He has published 90 international journal papers and more than 70 international conference papers. He is a CI for 18 competitive national grants. He serves/ served as an associate editor for four journals.

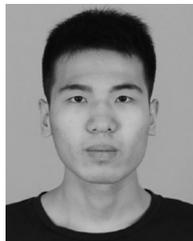

**Jiaye Li** is currently working toward the PhD degree with Central South University, China. His research interests include machine learning, data mining, and deep learning.

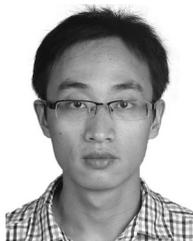

**Yanding Li** received the PhD degree in information and communication engineering from Xidian University, Xi'an, China. He is currently an associate professor with the Hunan Normal University, Changsha, China. His current research interests include medical image processing, machine learning, and data mining.

▷ **For more information on this or any other computing topic, please visit our Digital Library at** www.computer.org/csdl.